\documentclass[conference]{IEEEtran}
\IEEEoverridecommandlockouts
\usepackage{cite}
\usepackage{amsmath,amssymb,amsfonts}
\usepackage{algorithmic}
\usepackage{graphicx}   
\usepackage{textcomp}
\usepackage{xcolor}
\usepackage{hyperref}
\usepackage{float}
\usepackage{array} 
\usepackage{booktabs}
\usepackage[ruled,vlined,linesnumbered]{algorithm2e}
\makeatletter
\newcommand{\algorithmfootnote}[2][\footnotesize]{%
  \let\old@algocf@finish\@algocf@finish
  \def\@algocf@finish{\old@algocf@finish
    \leavevmode\rlap{\begin{minipage}{\linewidth}
    #1#2
    \end{minipage}}%
  }%
}
\makeatother\usepackage[tablename=TABLE]{caption}
\definecolor{codegreen}{rgb}{0.0,0.6,0.0}

\usepackage{subfigure}
\def\BibTeX{{\rm B\kern-.05em{\sc i\kern-.025em b}\kern-.08em
    T\kern-.1667em\lower.7ex\hbox{E}\kern-.125emX}}
\begin{document}

\title{IKDP : \\ Inverse Kinematics through Diffusion Process}

\author{\IEEEauthorblockN{Hao-Tang Tsui*}
\IEEEauthorblockA{\textit{College of ECE} \\
\textit{
\small{National Yang Ming Chiao Tung University}
}\\
henrytsui000@gmail.com}
\and
\IEEEauthorblockN{Yu-Rou Tuan*}
\IEEEauthorblockA{\textit{College of ECE} \\
\textit{
\small{National Yang Ming Chiao Tung University}
}\\
yztuan1129@gmail.com}
\and
\IEEEauthorblockN{Hong-Han Shuai}
\IEEEauthorblockA{\textit{College of ECE} \\
\textit{
\small{National Yang Ming Chiao Tung University}
}\\
hhshuai@nycu.edu.tw}

}

\maketitle

\begin{abstract}
It is a common problem in robotics to specify the position of each joint of the robot so that the end point reaches a certain target in space. This can be solved in two ways: 
\par 1. Forward kinematics method (FK) 
\par 2. Inverse kinematics method (IK)
\par However, inverse kinematics cannot be solved by an algorithm. The common method is the Jacobian inverse technique, and some people have tried to find the answer by machine learning. And in this project, we will show how to use the Conditional Denoising Diffusion Probabilistic Model (DDPM) \cite{2006.11239} to integrate the solution of calculating IK.
\\
\end{abstract}

\begin{IEEEkeywords}
Inverse kinematics, Denoising Diffusion Probabilistic Model, self-Attention, Transformer
\end{IEEEkeywords}

\section{Introduction}

\huge{T}\normalsize{he}
 motivation for doing this topic comes from our side project. We found that the inverse kinematics method(IK), which is commonly used to control the movement of human models in game engines, has a lot to be improved, such as the fidelity of movement and the angle of joint rotation accuracy. Therefore, we decided to do research on it.
\par Inverse kinematics is used to determine the kinematic equations to motion of a robot to reach a desired position. For example, given a 3D model of a human body, if the goal is to change the hand from a relaxed position to a waving position, we can use inverse kinematics to find the angle of the wrist and elbow.
\par Therefore, we hope to convert the IK problem into an objective function in machine learning, combined with the denoising diffusion probabilistic models(DDPM)\cite{2006.11239} taught this semester, consider the target condition that will be needed in the IK process, so our topic will use conditional DDPM to find IK's solution.
\par That is to train a model F with given target position $(t_x, t_y)$, so that $F(t_x, t_y)$ has the best solution ${\theta}$, ${\theta}$ is a series of robot joint angles, so that the end of the robot can reach $(t_x, t_y)$. Because we expect the robot's hand to reach a specific target, the target $(t_x, t_y)$ is added to the time embedding and becomes a condition diffusion form.
\begin{figure}[ht]
\includegraphics[width=\linewidth]{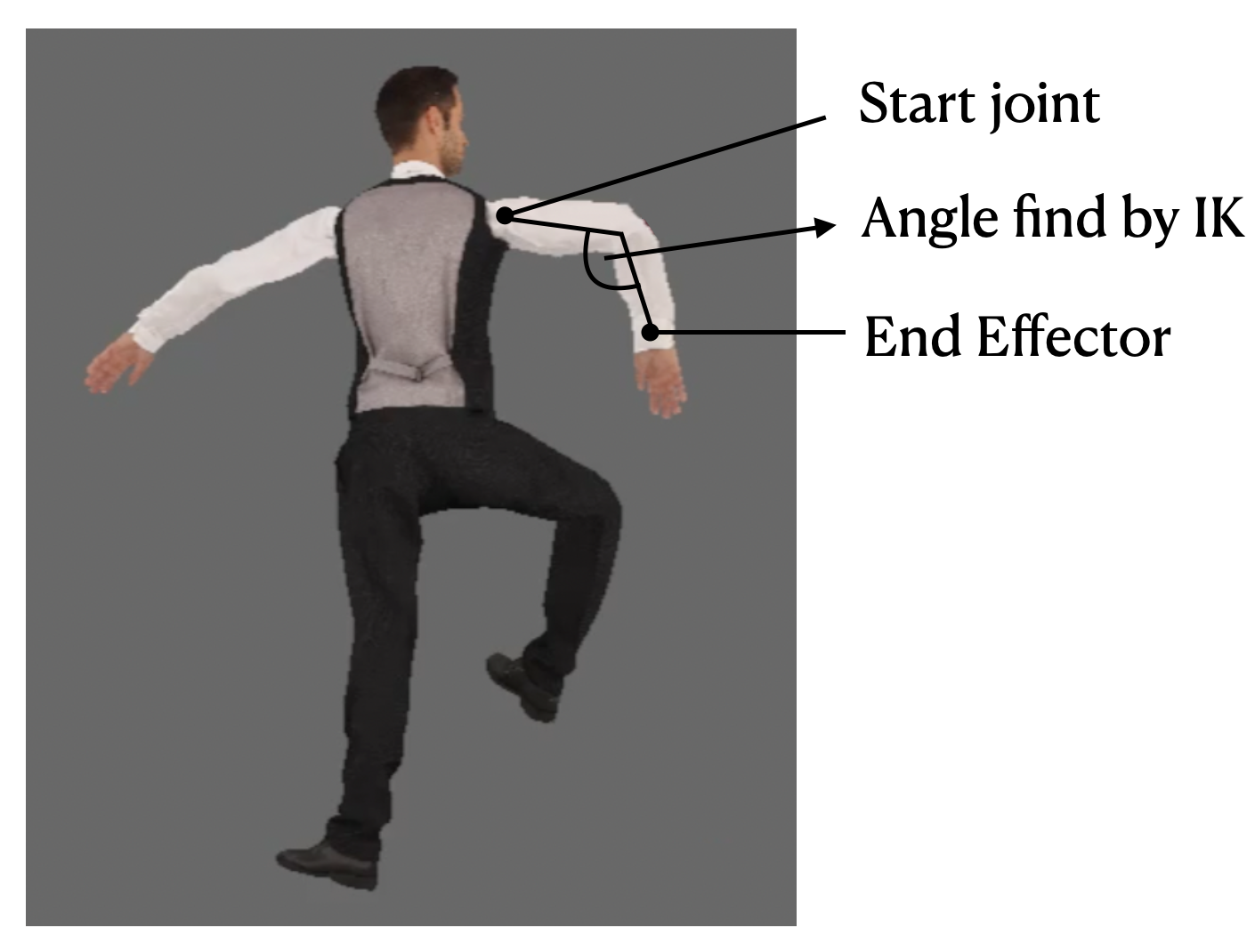}
\caption{Pose Estimation Calculation for the depth of joints}
\label{avatar}
\end{figure}

\section{Related Work}

\subsection{Inverse Kinematics Method (IK)}
Inverse kinematics method (IK) is responsible for calculating the joint movement required for the end position. It will calculate the angle of each joint from a terminal joint and the length of each bone to form a robotic arm that touches this distal point. At present, IK can be realized through quite a variety of algorithms, such as Jacobian method\cite{1401.1488} and machine learning methods that have been gradually proposed in recent years\cite{2205.10837}. To sum up, we think that IK is worth improving, so we plan to use machine learning to improve it.

\subsection{Denoising Diffusion Probabilistic Models (DDPM)}
Denoising Diffusion Probabilistic Models (DDPM), as mentioned in the class, continuously removes the noise from a gaussian noise through the encoder and decoder until it becomes the image we want to generate. DDPM can also be done in 1D, which is our topic's task. We would like to use this technology because the task of IK is also related to generation. We want to achieve the task of generating the angle of the joint point of the manipulator through Conditional DDPM\cite{2108.02938}.

\section{Methods}
\subsection{Ideas}
Design a diffusion model $F = ({t_x}, {t_y}, {t_z})$, given the target position ($t_x$, $t_y$, $t_z$) of the distal joint ${t}$ and the direction of the solution bone vector (two adjacent joints form a bone vector). 
\par
\begin{figure}[ht]
\includegraphics[width=\linewidth]{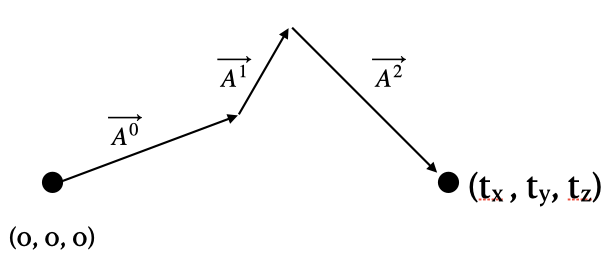}
\caption{Robotic arm vector schematic diagram}
\label{org}
\end{figure}

 With known bone length, we define that there are total ${N}$ joints on a robotic arm, so there will be ${N}$ bone vectors. 
Finally, the bone vectors will be combined into a vector list $ \vec{A} = \{{\vec{A^0}, \vec{A^1}, ...,\vec{A^{N-1}}} \}$ of the best solution, and the end of the robotic arm can reach target position $({t_x}, {t_y}, {t_z})$, shown as Fig. \ref{org}.
In three-dimensional space, each bone vector $\vec{A^n}$ has three components x, y, and z. We define that $\vec{A^n_x}$ is the x component of the $n_{th}$ bone, y and z components can be deduced by analogy. See equation (\ref{eq1}).
\begin{equation}\label{eq1}
t = 
\begin{bmatrix} t_x \\t_y \\t_z\end{bmatrix} = \sum_{n=0}^{N-1}\vec{A^n} = \sum_{n=0}^{N-1}\begin{bmatrix}\vec{A^n_x} \\
\vec{A^n_y} \\\vec{A^n_z}\end{bmatrix}
\end{equation}
\par
\begin{figure}[ht]
\includegraphics[width=\linewidth]{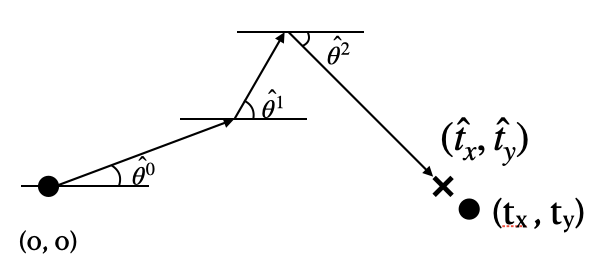}
\caption{Pose Estimation Calculation for the depth of joints}
\label{withcor}
\end{figure}

 We found that instead of calculating these bone vectors from position directly, computing the angular ratio vectors of the joints simplifies the problem. Which means to calculate the angle between the bone vectors and the standard coordinate system, because the length of these bone vectors is fixed. See equation (\ref{eq2}) and Fig. \ref{withcor}, $\theta^n$ represents the angle between the $n_{th}$ joint and the ${+x-}$axis, and $b^n$ represents the length of the ${n_{th}}$ bone vector. The sum of the cumulative rotations of the x, y, and z axes of each joint will be the position $({t_x}, {t_y}, {t_z})$ of the robotic arm's target ${t}$. 
\begin{equation}\label{eq2}
\scalebox{0.625}{%
t = $\sum_{n=0}^{N-1}\begin{bmatrix}
b^n \\
0 \\
0
\end{bmatrix}
\begin{bmatrix}
\cos{\theta^n_x} & 0 & \sin{\theta^n_y}\\
0 & 1 & 0 \\
-\sin{\theta^n_x} & 0 & \cos{\theta^n_x} 
\end{bmatrix}
\begin{bmatrix}
1 & 0 & 0 \\
0 & \cos{\theta^n_x} & -\sin{\theta^n_x}\\
0 & \sin{\theta^n_x} & \cos{\theta^n_x} 
\end{bmatrix}
\begin{bmatrix}
\cos{\theta^n_z} & -\sin{\theta^n_z} & 0\\
\sin{\theta^n_z} & \cos{\theta^n_z} & 0\\
0 & 0 & 1 
\end{bmatrix}$}
\end{equation}
\par Then we simplify the problem to only two dimensions. The position of target ${t}$ is $({t_x}, {t_y})$, and each bone length has a length of 1, shown as equation (\ref{eq3}). Our task becomes to design a model ${F}$ so that $F({t_x}, {t_y})$ obtains best solution $\hat{\theta} = \{\hat{\theta^0}, \hat{\theta^1}, ..., \hat{\theta^{N-1}} \}$. See below equation (\ref{eq4}), after substituting the bone vector of ${(1, 0)}$, it shows that ${t_x}$ in the target position is the sum of cosine values of each joint points' rotation angles, and ${t_y}$ is the sum of their sine value. 
 $(hat{t_x}, hat{t_y})$, that is the end joint that obtained by the model, can be calculated by ${\hat{\theta^n}}$ in the same way.
If we want to go back to the original question of 3-dimension, we only need to add another dimension from the sum to the rotated length of the joint angle and the inner product of the bone length.
\begin{equation}\label{eq3}
t = 
\begin{bmatrix}
t_x \\
t_y
\end{bmatrix} = 
\sum_{n=0}^{N}\begin{bmatrix}
1 \\
0
\end{bmatrix}
\begin{bmatrix}
\cos{\theta^n} & \sin{\theta^n}\\
-\sin{\theta^n} & \cos{\theta^n} 
\end{bmatrix}
\end{equation}
\begin{equation}\label{eq4}
t_x  = 
\sum_{n=0}^{N}
\cos{\theta^n}, \quad
t_y  = 
\sum_{n=0}^{N}
\sin{\theta^n}
\end{equation}
\begin{equation}\label{eq5}
\hat{t_x}  = 
\sum_{n=0}^{N}
\cos{\hat{\theta^n}}, \quad
\hat{t_y}  = 
\sum_{n=0}^{N}
\sin{\hat{\theta^n}}
\end{equation}

\subsection{Propose}
Train a Conditional DDPM model to gradually denoise from $\theta_T$ to $\hat{\theta_0}$. Where $\theta_T=\{\theta | N(\mu,\sigma)\}$ represents a list of noise sample from Gaussian, and $\hat{\theta_0} = \{ {\hat{\theta_0^0}, \hat{\theta_0^1}, ... \hat{\theta_0^{N-1}} } \}$ represent the angle to be rotated by each joint. $\hat{\theta_0}$ should be close to the ground truth $\theta_0$, make $\hat{t} = (\hat{t_x}, \hat{t_y})$ close to ground truth $t = (t_x, t_y)$. 
\par In Inverse Kinematics task, it is not enough to use only the diffusion model, because we also need to turn the tip to the target, so we will add condition ${t_x}$, ${t_y}$ to indicate where the tip of the robot arm should point. That is, we hope that $\hat{t}$ is close to ${t}$.

\begin{figure}[ht]
\includegraphics[width=\linewidth]{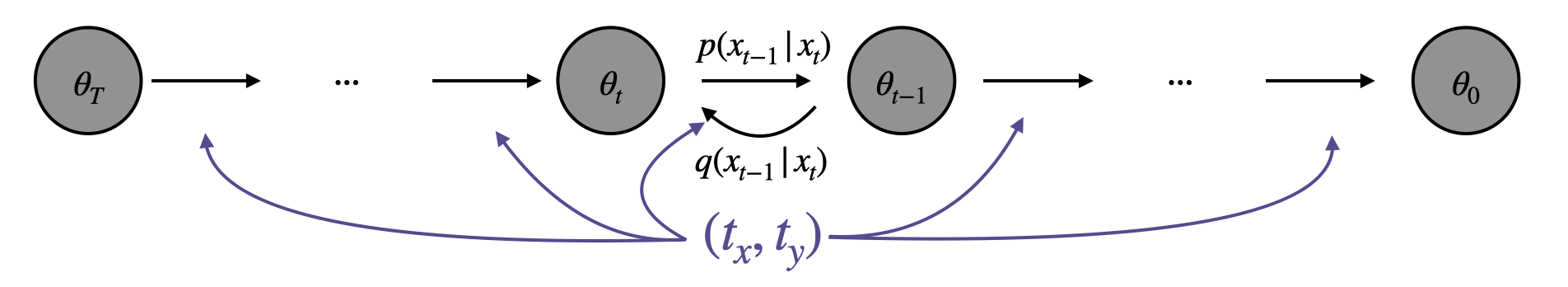}
\caption{Process of training conditional DDPM model}
\label{process}
\end{figure}

\subsection{Loss Function}

Evaluating the performance of Inverse Kinematics method will need to calculate the distance or difference, so we think Mean-Square Error(MSE) will be a good choice. In our model, there are a total of two losses as indicators for evaluate the performance of the model, and we expect both losses are as low as possible. The two matrices we evaluated, namely:
\subsubsection{Distance of ${t}$}
$||t-\hat{t}||_2 = \sqrt{(t_x-\hat{t_y})^2+(t_y-\hat{t_y})^2}$
\\Evaluates the quality of the model, and represents the differences of the end joint of the robotic arm and target position.
\subsubsection{Loss of ${\theta}$}
$||\theta - \hat{\theta}||_2 = \sqrt{\Sigma_{n=0}^{N-1}{(\theta^n-\hat{\theta^n})^2}}$
\\Represents the difference between each joint point and ground truth, and it is the loss function which would do backward.

\subsection{Model Architecture}
\begin{figure}[ht]
\includegraphics[width=\linewidth]{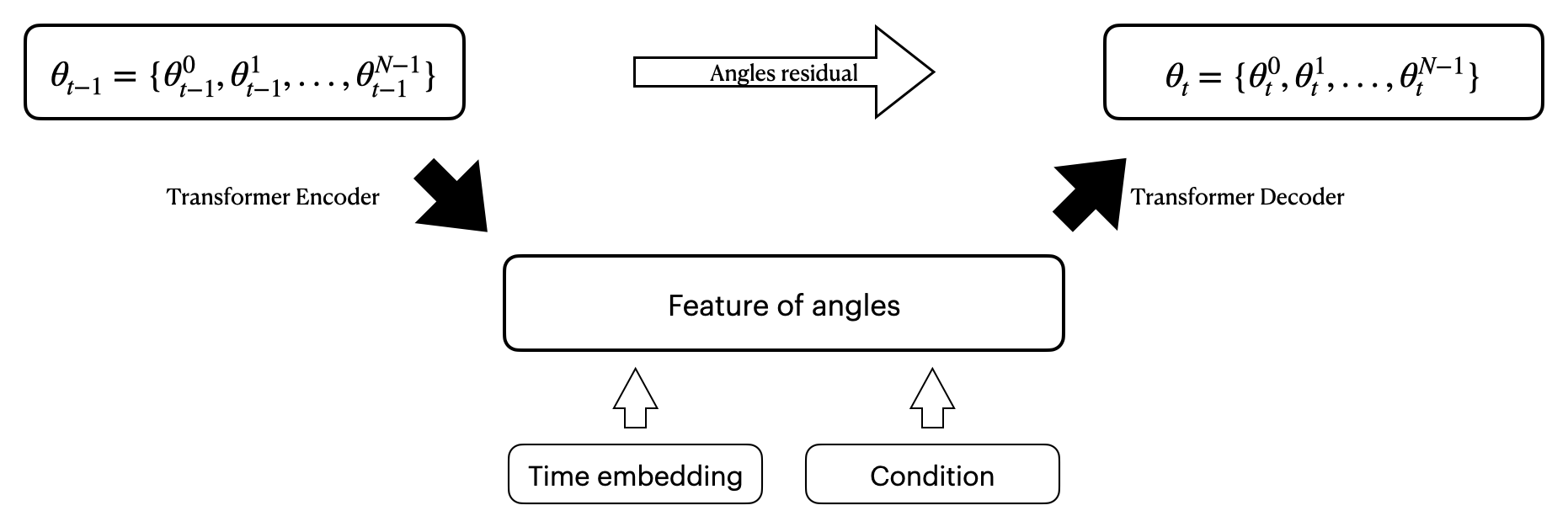}
\caption{Pose Estimation Calculation for the depth of joints}
\label{model}
\end{figure}

In a timestep, we hope that the input to the model is $\theta_{t-1} = \{\theta^0_{t-1}, \theta^1_{t-1}, ..., \theta^{N-1}_{t-1}\}$, which is a list indicated how many degrees each joint of the robotic arm should turn. After input through a transformer encoder, we can obtain the feature. Concate the feature with the time embedding and condition which is the target position $(t_x, t_y)$. Its output will pass through a few layers of MLP, finally, go through the transformer decoder, and add the residual of the previous $\theta_{t-1}$ to become the $\theta_{t}$ we want, that is, the angle of each joint after 1 timestep of denoising process. The model architecture diagram is Fig. \ref{model}

\subsection{Training Process}

\begin{algorithm}[!h]
\SetAlgoLined
\DontPrintSemicolon
\SetNoFillComment
\footnotesize
\KwIn{A list of ground truth $\theta$}

Initialization: ${t_x, t_y} \leftarrow \sum^{N-1}_{n=0}{\cos{\theta^n}}, \sum^{N-1}_{n=0}{\sin{\theta^n}}$\;
$t \sim Uniform({1, ..., T})$\;
$\epsilon \sim \mathcal{N}(0, I)$ \;
$\hat{\theta} \leftarrow model(\sqrt{\hat{\alpha}} \theta_0 + \sqrt{1-\hat{\alpha_t}}\epsilon, (t_x, t_y), t)$ \;
$Loss \leftarrow MSE(\hat{\theta}, \theta)$ \;
$Dist \leftarrow MSE(\sum^{N-1}_{n=0}{\cos{\hat{\theta^n}}}, \sum^{N-1}_{n=0}{\sin{\hat{\theta^n}}})$ \;
\caption{Pseudo-code of training IKDP.}
\label{algo:byte}
\end{algorithm}

\begin{algorithm}[!h]
\SetAlgoLined
\DontPrintSemicolon
\SetNoFillComment
\footnotesize
\KwIn{A target pair $t_x, t_y$}
\KwOut{The prediction $\hat{\theta}$}

Initialization: $t \sim Uniform({1, ..., T})$\;
\For{$t$ in [T, ..., 1]}{
    $z \sim \mathcal{N}(0, I)$ if $t > 1$, else $z = 0$ \;
    $\theta_{t-1} \leftarrow \frac{1}{\sqrt{\alpha_t}}(\theta_t-\frac{1-\alpha_t}{\sqrt{1-\alpha_t}}\epsilon(\theta_t, t))+\sigma_tz$ \;
}
Return: $\theta_0$
\caption{Pseudo-code of sample IKDP.}
\label{algo:byte}
\end{algorithm}

\section{Experiments}

\subsection{Visualization}
\begin{figure}[t]
\includegraphics[width=\linewidth]{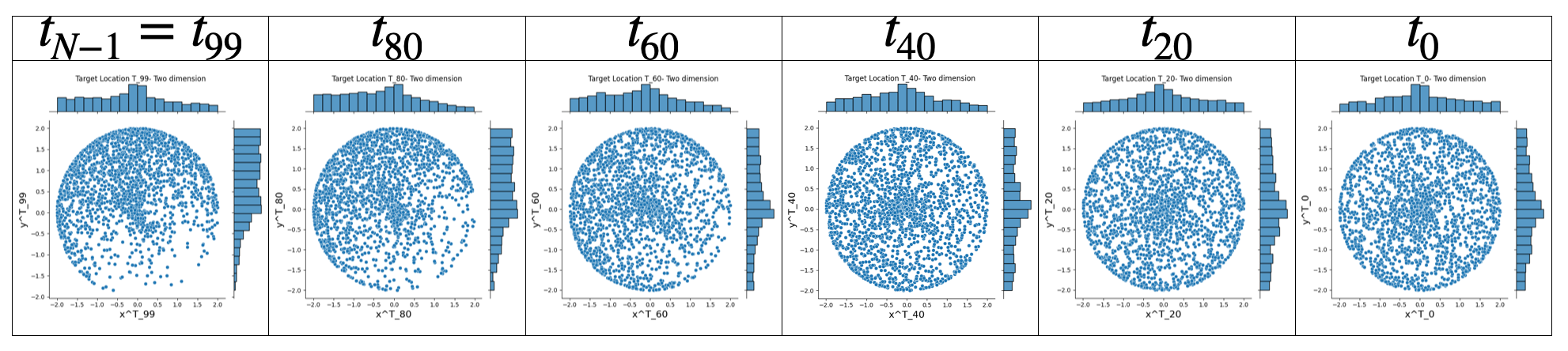}
\caption{Unity distribution in different diffusion time steps}
\label{gaussian}
\end{figure}

We have visualized the training process and data sets. The first is to gradually add noise following the Gaussian distribution to the target point. Although in the normal Cartesian coordinate system, the trend of the Gaussian distribution that he gradually becomes cannot be observed. But you can map it back to the $\arccos{t_x}$ coordinate system, and you will find that it will gradually change from a uniform distribution (because I drew tens of thousands of target coordinates at a time) to a normal distribution. Please see the Fig. \ref{gaussian} from the uniform distribution at $t_0$ to the normal distribution.

\begin{figure}[t]
\centering
\subfigure[Training distance of IKDP]{
\label{dist}
\includegraphics[width=0.20\textwidth]{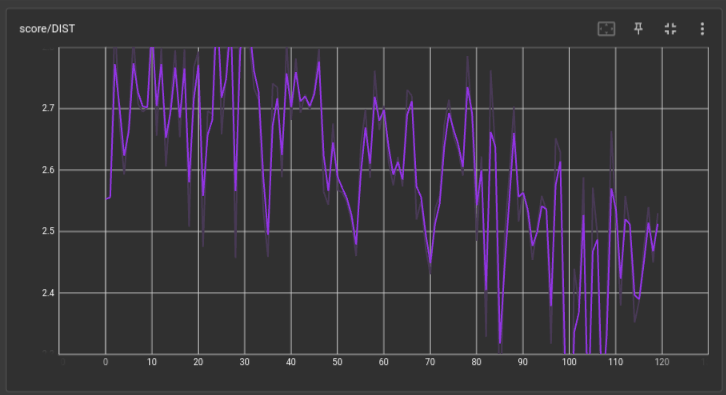}}
\subfigure[Training loss of IKDP]{
\label{loss}
\includegraphics[width=0.20\textwidth]{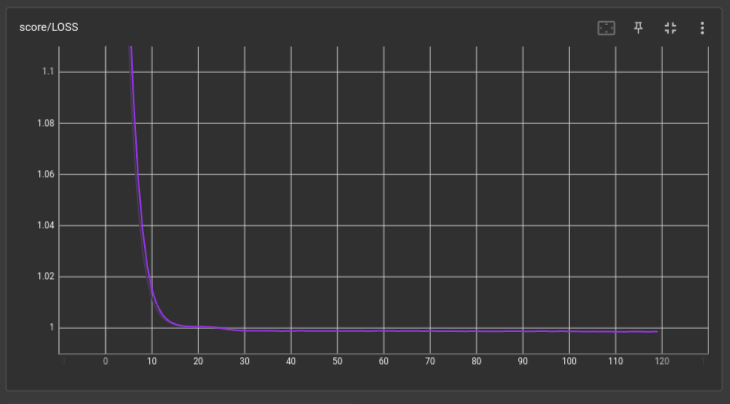}}
\caption{In the training process of our IKDP, it can be seen that the Loss aspect converges quickly, and then slowly decreases; while the distance part gradually decreases like training FID, but it does not strictly decrease.}
\label{train_loss}
\end{figure}
In the training part, we compare the output by the model with the correct angle $(\hat{\theta}, \theta)$, and visualize the distance between the tip and the actual target after the angle is applied to the robot $(\hat{t}, t)$. It can be found that the latter converges quickly because it is used as a loss function for model training. The former is positively correlated with loss, that is, there is a tendency to gradually decrease, just like training an ordinary Diffusion process, FID will not strictly decrease, but the trend will gradually decrease. In the Fig. \ref{train_loss}, you can see the values recorded by tensorboard.

\begin{figure}[t]
\includegraphics[width=\linewidth]{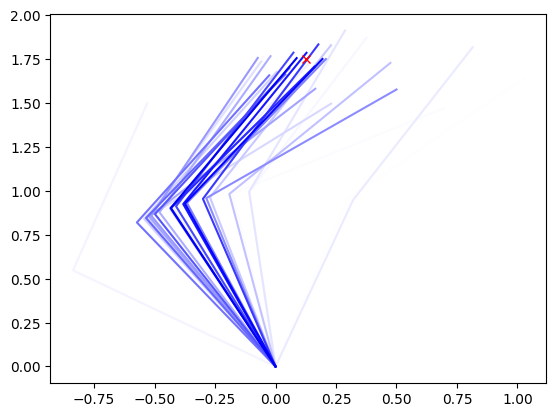}
\caption{Pose Estimation Calculation for the depth of joints}
\label{vis}
\end{figure}

\par Next is to visualize the diffusion process of a certain target. In the figure Fig. \ref{vis}, the red cross is the given target $(t_x, t_y)$, and each line is the denoise process. What is relatively light is the joint angle combination that has not been denoised. After several rounds of generation, it will gradually become darker, and the darkest is the result of the final diffusion, that is, the generated joint ends $(\hat{t_x}, \hat{t_y})$ correctly point to the target $(t_x, t_y)$.
\\

After designing the model and visualization, we did several experiments.

\begin{table}[t]
\begin{center}
\setlength{\tabcolsep}{2.5pt}

\begin{tabular}{ l | c c}

\toprule
Timesteps & Angle Distance$\downarrow$ & Target Distance$\downarrow$\\
\midrule

2 & 0.499 & 1.982 \\
4  & \textbf{0.32} & \textbf{1.283} \\
8 & 0.243 & 8.03 \\
\bottomrule
\end{tabular}
\end{center}
\caption{For experiments with different joints, we speculate that the more joints the higher the fault tolerance rate. In the experiment of two joint points, there is almost only one set of inverse kinematics solutions for each angle, so the error of one joint angle can easily lead to a long distance from the target.}
\label{table_joint}
\end{table}

\subsection{Experiments with different joints}
The first is the experiment on the number of different joint points. We did experiments with 2, 4, and 8 joints respectively. It can be found that the more joints, the smaller the loss. For such a result, we speculate that the more joints the higher the fault tolerance rate. In the experiment of two joint points, there is almost only one set of inverse kinematics solutions for each angle, so the error of one joint angle can easily lead to a long distance from the target.

\begin{table}[t]
\begin{center}
\setlength{\tabcolsep}{2.5pt}

\begin{tabular}{ l | c c}

\toprule
Timesteps & Angle Distance$\downarrow$ & Target Distance$\downarrow$\\
\midrule

40 & 1.382 & 4.361\\
80 & \textbf{1.283} & 3.611\\
100 & 1.37 & \textbf{3.44} \\
\bottomrule
\end{tabular}
\end{center}
\caption{For experiments with different timesteps, it can be found that 80 timesteps are more satisfactory in performance, and larger or smaller sizes will not have better preformance.}
\label{table_timestep}
\end{table}

\subsection{Experiments at different timesteps}
We think that in this task, unlike DDPM for image generation, the task of generating one-dimensional numbers does not require so many diffusion processes. On the one hand, it can speed up the generation and save unnecessary iterations. After adjusting the parameter settings many times, it can be seen from the experimental results that the highest accuracy rate will be achieved at 80 timesteps, which also verifies the previous speculation. The results are shown in Table \ref{table_timestep}.

\begin{table}[t]
\begin{center}
\setlength{\tabcolsep}{2.5pt}

\begin{tabular}{ l | c c}

\toprule
Methods & Target Distance$\downarrow$ & Runtimes$\downarrow$\\
\midrule

MLP & 12.3 & 0.04\\
IKNET \cite{2205.10837} & 2.3 & \textbf{0.007}\\
\textbf{IKDP (Ours)} & \textbf{1.98} & 0.065 \\
\bottomrule
\end{tabular}
\end{center}
\caption{Compared with the previous state-of-the-art that used machine learning to find the inverse kinematics solution, it can be seen that the accuracy is much higher than the former, but the execution speed is slightly slower.}
\label{table_ablation}
\end{table}

\subsection{Compare with other models}
Although this topic is not popular, there are indeed other teams doing research here. See the Table\ref{table_ablation}. Like IKNET\cite{2205.10837} proposed in early 2022, or simply MLP. Therefore, we are here to make some comparisons with other people's work, but because of the different methods, they only provide the mean-square error of the distance from the target, without the distance between each joint, so they only compare the former with the execution time. Our work is much better than other machine learning methods, such as IKNET. In terms of execution time, DDPM\cite{2006.11239} is slower than pure MLP or IKNET because it takes many iterations to calculate the answer. However, like the nature of DDPM, it has a relatively high generation quality.

\subsection{Future Work}
In addition to the currently completed content, we are also advancing our project.
First of all, we think that the model still has a lot of information to consider, and only the Transformer Encoder/decoder seems to be a bit insufficient.
In addition, as mentioned above, this technology can be actually applied to joint control to determine whether the theoretically feasible model will have serious delay or whether it is accurate.
Finally, on the topic of inverse kinematics, in fact, in addition to expecting the end of the robot to reach the given coordinates (x, y, z), it is also hoped that the contact angle of the robot will reach (yaw, pitch, roll). We believe that the model plus other design can accomplish this task.

\section{Conclusion}
In this Final project, we use the DDPM\cite{2006.11239} concept to calculate the Inverse kinematic, that is, to generate the angles that each joint of the robot should have under the given target position of the robot tip.
At the same time, we changed the ordinary DDPM into a conditional DDPM by adding conditions at the embedding time, so that the model can generate the joints of the robot so that the end touches our given conditions. In terms of generation ability, it is much better than the previous SOTA (IKNET), and the generation speed is also within one second.
Finally, through the simplification of the formula, we observe the rotation matrix between the bones and change the formula that was originally calculated separately into an inner product or a parallel addition. The original data generation speed is increased to the millisecond level, and the data set of 100,000 records can be quickly generated.

\section*{Acknowledgment}

Here we are very grateful to the contributors in various fields, from the developers of the virtual game engine, for finding the research topic among them; and the Denoising Diffusion Probabilistic Models in image generation and the condition DDPM derived by other teams are also Inspired our design model. Of course, the professors and teaching assistants of this course also gave us valuable opinions and ideas.


\begin{thebibliography}{00}
\bibitem{2006.11239}
Jonathan Ho, Ajay Jain and Pieter Abbeel.
\newblock Denoising Diffusion Probabilistic Models, 2020;
\newblock arXiv:2006.11239.

\bibitem{2205.10837}
Raphael Bensadoun, Shir Gur, Nitsan Blau, Tom Shenkar and Lior Wolf.
\newblock Neural Inverse Kinematics, 2022;
\newblock arXiv:2205.10837.

\bibitem{1401.1488}
Zeeshan Bhatti, Asadullah Shah, Farruh Shahidi and Mostafa Karbasi.
\newblock Forward and Inverse Kinematics Seamless Matching Using Jacobian, 2014,
\newblock Sindh University Research Journal (SURJ) Volume 45 (2), 8/2013,
  pp:387-392, Sindh University Press;
\newblock arXiv:1401.1488.

\bibitem{2108.02938}
Jooyoung Choi, Sungwon Kim, Yonghyun Jeong, Youngjune Gwon and Sungroh Yoon.
\newblock ILVR: Conditioning Method for Denoising Diffusion Probabilistic Models, 2021;
\newblock arXiv:2108.02938.

\end{thebibliography}
\end{document}